\documentclass{article}

\usepackage[final,nonatbib]{tackling_climate_workshop_style}

\usepackage[utf8]{inputenc}
\usepackage[T1]{fontenc}
\usepackage{hyperref}
\usepackage{url}
\usepackage{booktabs}
\usepackage{amsfonts}
\usepackage{amsmath}
\usepackage{nicefrac}
\usepackage{microtype}
\usepackage{graphicx}
\usepackage{subcaption}

\usepackage{comment}

\title{Zero-Shot Forecasting of In-Situ Soil Moisture and Temperature with a Time-Series Foundation Model}

\author{%
  Mohammad Talaei$^{1}$ \quad
  Mohammad Goudarzi$^{1}$ \quad
  Maille Todd$^{2}$ \\
  \textbf{Vanessa N.L. Wong}$^{2}$ \quad
  \textbf{Iadine Chad\`{e}s}$^{3}$ \\
  $^{1}$Faculty of Information Technology, Monash University, Australia \\
  $^{2}$School of Earth, Atmosphere and Environment, Monash University, Australia \\
  $^{3}$The Environmental Informatics Hub, Data Science and AI, Monash University, Australia \\
  \texttt{\{mohammad.talaei, mohammad.goudarzi, maille.todd1\}@monash.edu} \\
  \texttt{\{vanessa.wong, iadine.chades\}@monash.edu}
}

\begin{document}

\maketitle

\begin{abstract}
Soil moisture and temperature govern plant-water availability and microbial activity, making them key indicators for evaluating land-restoration efforts on degraded landscapes, but the physical and data-driven models normally used to forecast them both need information that a newly instrumented restoration site does not yet have: hydraulic parameters for the former, months of per-site training data for the latter. We evaluate \textbf{TimesFM}\footnote{\url{https://huggingface.co/google/timesfm-2.0-500m-pytorch}}, a 500M-parameter time-series foundation model, \emph{zero-shot} on 15-minute telemetry from 27 sensor stations across three land-management regimes (Control, Treatment, Scald) at a semi-arid restoration site, forecasting 5\,cm and 30\,cm soil moisture and enclosure temperature at seven lead times from 1\,h to 168\,h. Using gap-aware rolling-origin evaluation across sensors, we find TimesFM beats the best rule-based statistical forecasting baselines (persistence, seasonal-naive, or 7-day climatology) by 21--76\% at sub-daily lead times but loses by 6--79\% beyond a day, a pattern explained by a diurnal-alignment mechanism that holds for both soil-moisture depths and temperature. A CatBoost baseline trained walk-forward on the same origins with full history and weather covariates never wins overall, and we show its weakness is partly a cold-start effect specific to short deployments. We additionally quantify the inference cost of zero-shot forecasting on commodity CPU hardware. Our results highlight a practical trade-off between zero-shot forecasting capability, predictive performance, and computational efficiency.
\end{abstract}

\section{Introduction}
\label{sec:intro}

Soil is a non-renewable resource underpinning food security, carbon storage, and biodiversity, yet land degradation through vegetation clearing, erosion, and altered hydrology threatens these services~[1,2]. The physical processes driving degradation and recovery are critical to understand in the plant root zone in the upper soil profile: soil moisture regulates plant-water availability and solute transport and is known to vary substantially with land-surface heterogeneity~[3], while soil temperature governs microbial and root activity. Forecasting these variables in situ is therefore useful for evaluating restoration interventions such as \emph{landscape rehydration}~[4], the use of soft structures (hay bales, mulch, brush packing) to slow runoff and rebuild degraded land surfaces.

Two practical obstacles stand in the way. Physically-based hydrological models~[5] need hydraulic parameters that are rarely measured at restoration sites. Data-driven forecasters (e.g., recurrent networks~[6,7]), meanwhile, are normally trained per site and per sensor, which is unworkable for a newly instrumented reserve with only a few months of telemetry. \emph{Time-series foundation models}, pre-trained across billions of heterogeneous time points, promise zero-shot forecasting without this per-site training step~[8,9,10,11,12]. We ask a direct question: on real, gap-ridden in-situ sensor data, does zero-shot forecasting with TimesFM, without any per-site training, actually hold up, and where does it break down? We examine not only predictive performance, but also the computational efficiency of inference, with particular emphasis on memory usage and inference time. Three key contributions are made: (1) A systematic evaluation of TimesFM zero-shot forecasting against statistical and tree-based baselines, considering both accuracy and inference efficiency. (2) Identification of distinct short- and long-horizon performance regimes linked to diurnal alignment across multiple environmental variables. (3) Demonstration that limited training history contributes to tree-model underperformance, highlighting trade-offs between adaptation, forecasting accuracy, and computational cost.

\section{Data and methods}
\label{sec:methods}

\begin{figure}[h]
  \centering
  \begin{subfigure}[b]{0.36\linewidth}
    \includegraphics[width=\linewidth]{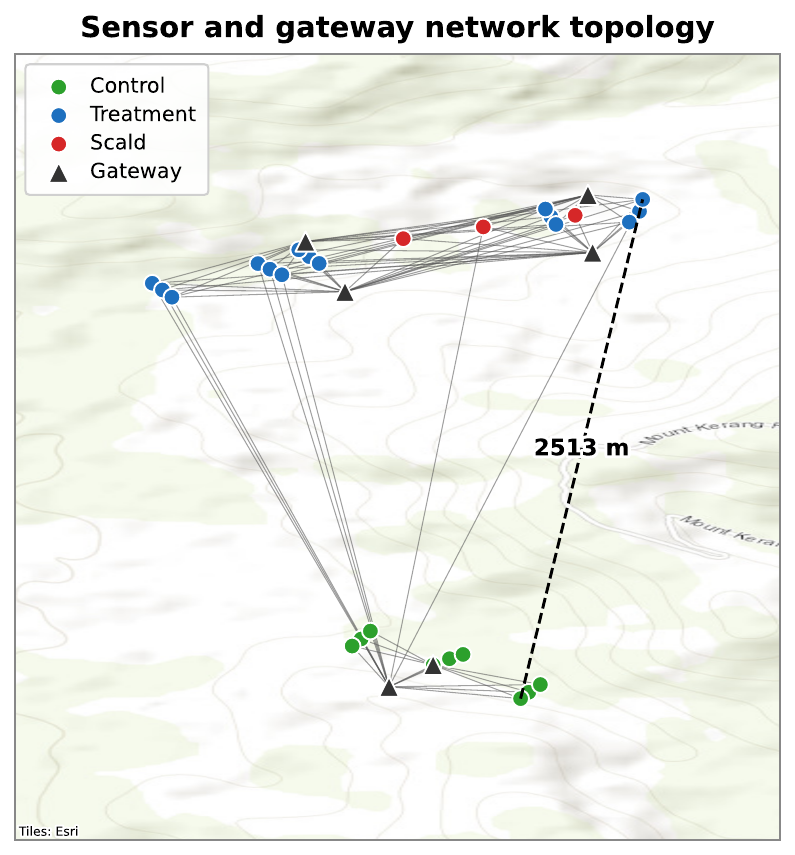}
    \caption{}
    \label{fig:topology-sub}
  \end{subfigure}
  \hfill
  \begin{subfigure}[b]{0.59\linewidth}
    \includegraphics[width=\linewidth]{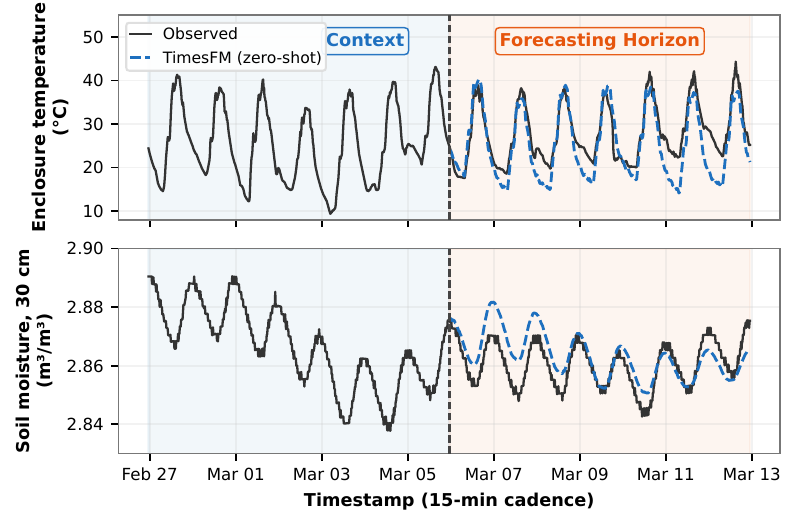}
    \caption{}
    \label{fig:forecast-sub}
  \end{subfigure}
  \caption{(a) Sensor–gateway topology, with the farthest sensor pair (2.5 km) marked; (b) a representative zero-shot forecast for a Treatment-regime sensor, showing 7 days of context and TimesFM’s 7-day-ahead forecast against ground truth for enclosure temperature and 30 cm soil moisture.}
  \label{fig:topology}
\end{figure}

\paragraph{Site and sensors.} Telemetry comes from Nardoo Hills Reserves, $\sim$210\,km north-west of Melbourne, Australia: 27 sensor stations (15 Treatment, 9 Control, 3 Scald) and 6 gateways at 15-minute cadence over a six-month deployment (Oct.\ 2024--Apr.\ 2025; Fig.~\ref{fig:topology}), spanning up to 2.5\,km. Control sites are undisturbed native vegetation; Treatment sites are under active landscape rehydration practices; Scald sites are severely degraded, bare ground with no intervention. We forecast soil moisture at depths of 5 and 30\,cm and logger \emph{enclosure} temperature. The latter represents temperature inside the sensor housing and is treated as a proxy for air temperature; it does not represent soil temperature at depth.

\paragraph{Gap-aware rolling-origin evaluation.} Raw series carry NaN gaps up to 69 days long; naively dropping them would splice non-contiguous timestamps and break the diurnal alignment a 15-minute-grid model relies on. We instead require every forecast origin (the timestamp from which a forecast is issued, using only data strictly before it) to sit inside a contiguous, fully-observed window covering both the longest context (2048 steps, $\approx$21 days) and the longest horizon (168\,h). Origins are chosen greedily, $\geq$7 days apart per sensor, giving 206 origins (18 sensors) for both moisture depths and 137 origins (16 sensors) for temperature, at seven lead times: 1, 4, 8, 16, 24, 96, and 168\,h.

\paragraph{Model.} TimesFM v2.0 (498.8M parameters, decoder-only transformer~[13], patch-based tokenization~[14], patch size 32) is run strictly zero-shot. As untrained, rule-based statistical baselines we use persistence (the value at the most recent timestep, held constant), seasonal-naive (the same 15-min slot 24\,h earlier), and 7-day climatology (the mean of the same clock time over the preceding week).

\paragraph{Tree-based baseline.} We add CatBoost~[15], the strongest of four gradient-boosting variants (XGBoost, LightGBM, CatBoost, HistGBR) studied in earlier model-selection experiments, trained on all history before each origin, using lagged time-series features of the target (rolling-window statistics at multiple look-backs) plus weather covariates, an intentionally generous comparison, since TimesFM sees only its own univariate history capped at 2048 steps. To avoid leaking the future, we retrain CatBoost at each 14-day cutoff using only feature and target timestamps that precede it, verified across all 518 fitted groups.

\section{Results}
\label{sec:results}

\begin{figure}[t]
  \centering
  \includegraphics[width=\linewidth]{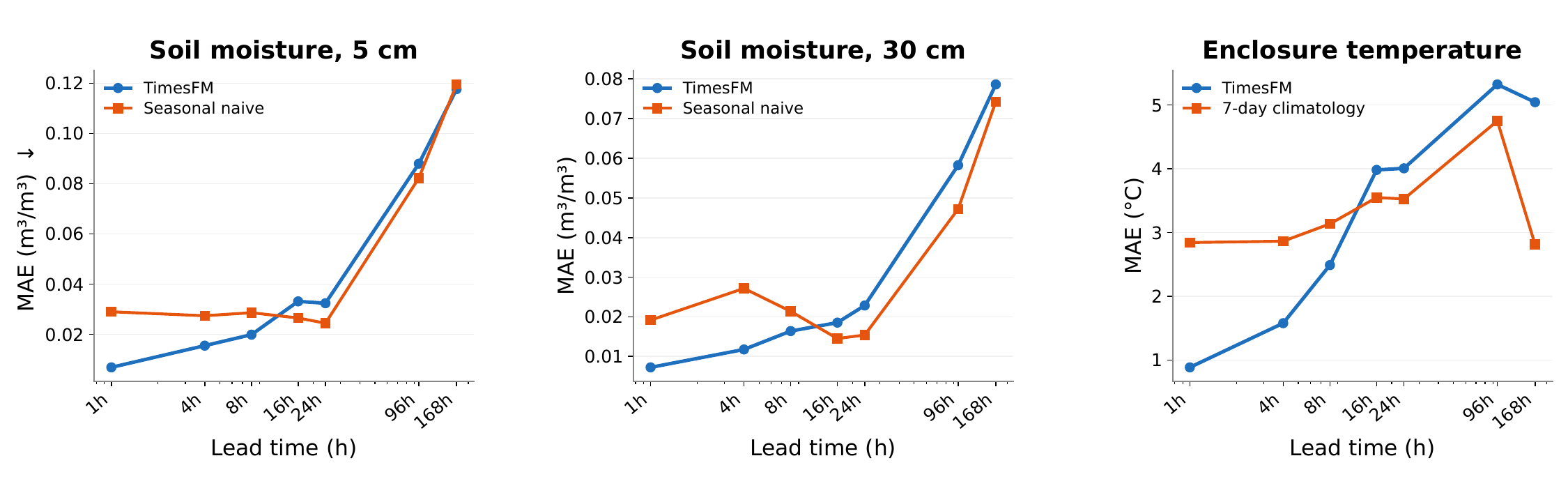}
  \caption{TimesFM vs.\ the best baseline (chosen once per target by lowest mean normalized MAE across all lead times, held fixed, not re-chosen per horizon), pooled across sensors.}
  \label{fig:headline}
\end{figure}

\textbf{TimesFM wins short-term, loses long-term.} Table~\ref{tab:baselines} and Fig.~\ref{fig:headline} show a consistent regime split across all three targets. At 1--8\,h, TimesFM beats the best classical baseline by 21--76\%; beyond 16\,h it loses by 6--79\%, with 7-day climatology overtaking everything past four days. This is not an artifact of comparing against a weak opponent: for moisture, seasonal-naive is the strongest baseline overall (persistence and climatology both do worse on average), and TimesFM still needs the sub-daily regime to win, consistent with broader findings that simple baselines are surprisingly hard for complex forecasters to beat~[16,17]. The crossover is explained by a diurnal-alignment mechanism: the ratio of 7-day peak-to-peak amplitude to 24-h drift is 4.1 (5\,cm), 3.4 (30\,cm), and 6.6 (enclosure temperature), largest for temperature and smallest for the deeper, more damped moisture signal, exactly as the mechanism predicts, and consistent with persistence's near-exact tie with TimesFM at 24\,h multiples in Table~\ref{tab:baselines} (moisture 24\,h and 168\,h rows).

\begin{table}[t]
\caption{MAE by method, target, and lead time ($\downarrow$ the better), pooled across sensors. Bold = the best.}
\label{tab:baselines}
\centering
\footnotesize
\begin{tabular}{lccccccccc}
\toprule
 & \multicolumn{3}{c}{5\,cm moisture (m$^3$/m$^3$) $\downarrow$} & \multicolumn{3}{c}{30\,cm moisture (m$^3$/m$^3$) $\downarrow$} & \multicolumn{3}{c}{Encl. temp.\ ($^\circ$C) $\downarrow$} \\
\cmidrule(lr){2-4}\cmidrule(lr){5-7}\cmidrule(lr){8-10}
Method & 1h & 24h & 168h & 1h & 24h & 168h & 1h & 24h & 168h \\
\midrule
TimesFM (zero-shot) & \textbf{0.0069} & 0.0324 & 0.1176 & \textbf{0.0072} & 0.0229 & 0.0786 & \textbf{0.89} & 4.01 & 5.04 \\
Persistence & 0.0101 & \textbf{0.0244} & 0.1195 & 0.0074 & \textbf{0.0154} & 0.0743 & 1.73 & 3.82 & 5.06 \\
Seasonal naive & 0.0290 & 0.0244 & 0.1195 & 0.0192 & 0.0154 & 0.0743 & 2.43 & 3.82 & 5.06 \\
7-day climatology & 0.0792 & 0.0669 & \textbf{0.0695} & 0.0579 & 0.0473 & \textbf{0.0467} & 2.84 & \textbf{3.53} & \textbf{2.81} \\
CatBoost & 0.0316 & 0.0815 & 0.1738 & 0.0448 & 0.0921 & 0.1719 & 1.40 & 4.01 & 5.04 \\
\bottomrule
\end{tabular}
\end{table}

\begin{figure}[t]
  \centering
  \includegraphics[width=\linewidth]{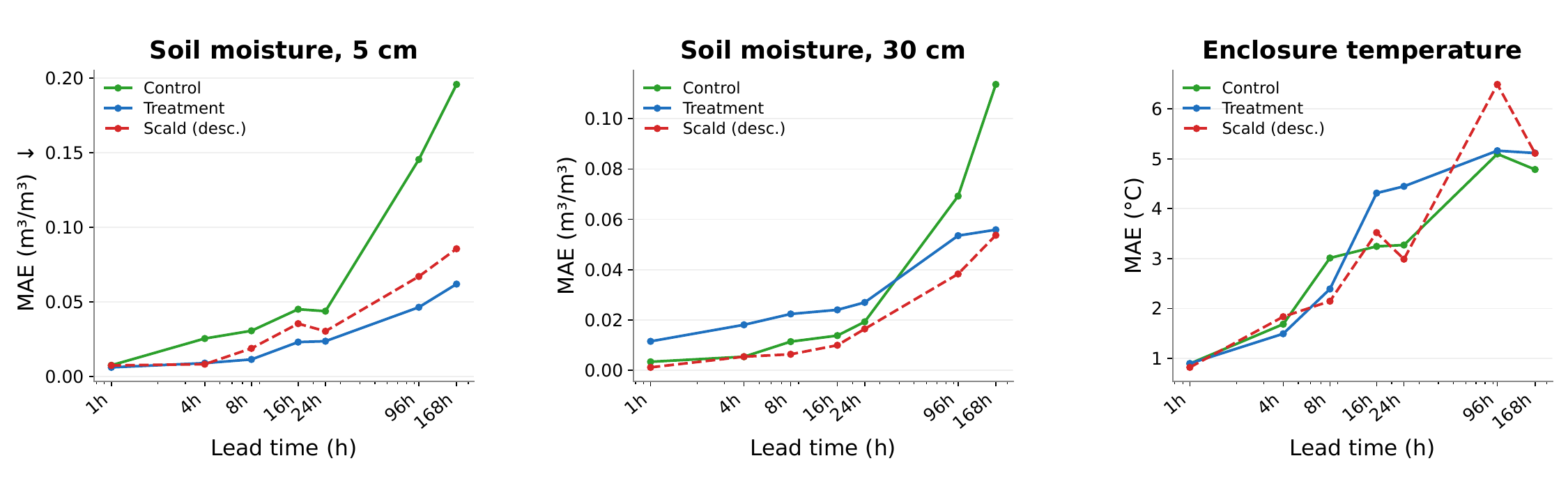}
  \caption{TimesFM MAE by land-management regime. Scald (3 sensors) is descriptive only.}
  \label{fig:exptype}
\end{figure}

\textbf{Behaviour is consistent across land-management regimes.} Fig.~\ref{fig:exptype} breaks TimesFM's MAE out by Control, Treatment, and Scald. The short/long regime split holds within every group; absolute error is generally higher for Control at longer moisture horizons, plausibly reflecting different local soil dynamics, though with only 3 Scald sensors that group is reported descriptively and its curve should not be over-interpreted.

\textbf{CatBoost's weakness is partly a cold-start effect.} Pooled across the season, CatBoost never wins (Table~\ref{tab:baselines}), despite seeing full history and weather covariates that TimesFM does not. We tested whether this reflects undertraining early in the record, when a walk-forward model has little history to fit on: error correlates negatively with training-set size (5\,cm moisture: $r=-0.25$), and MAE is 3.3$\times$ worse in the smallest-training-data quartile than the largest. Restricting to only the best-trained half of origins narrows the gap but does not close it. CatBoost remains weakest or second-weakest for all three targets even with a fair, data-rich shot.

\begin{table}[t]
\caption{System performance, measured locally. CatBoost retrains at every 14-day cutoff and per horizon (518 refits total); memory is the incremental cost of one fit/inference call with data already resident. Rule-based baselines are omitted; their cost is negligible by construction.}
\label{tab:sysperf}
\centering
\footnotesize
\begin{tabular}{lccccc}
\toprule
Method & Parameters & Training time & Inference time & Points/call & Peak memory \\
\midrule
TimesFM (zero-shot) & 498.8M & N/A & 6.4\,s & 672 & +2.7\,GB \\
CatBoost            & --     & 5.2\,s/refit & 0.003\,s & 1 & $<$5\,MB \\
\bottomrule
\end{tabular}
\end{table}

\textbf{Inference cost.} Table~\ref{tab:sysperf} compares the performance of zero-shot TimesFM and CatBoost on commodity hardware (Apple MacBook Pro, M4 CPU, 24 GB RAM). As shown in Table~\ref{tab:baselines}, TimesFM consistently outperforms CatBoost across all experiments; however, this advantage comes with higher inference time (6.4\,s) and peak memory usage (+2.7\,GB). Notably, TimesFM requires no training and can be deployed off the shelf, whereas CatBoost incurs a modest training overhead.

\section{Discussion, limitations, and future work}
\label{sec:limitations}
Overall, time-series foundation models (specifically TimesFM used in this study) offer a compelling cold-start solution for newly instrumented sites, enabling useful sub-daily forecasts without site-specific training or GPU infrastructure. This flexibility, however, comes with higher inference-time and memory costs, making foundation-model deployment an important design trade-off. In our experiments, this trade-off favours TimesFM at shorter horizons, while seasonal and climatological baselines remain more effective beyond one day for soil moisture and temperature.

Our conclusions are limited to six months of observations from a single pilot site. They therefore do not establish that the observed short- and long-horizon performance patterns will persist across seasons, climates or sensor networks. The Scald group has only 3 sensors, so its results are descriptive, not inferential. ``Enclosure temperature'' is an air-temperature proxy, not soil temperature at depth. Inference was CPU-only; a GPU deployment would be faster. Moisture units follow raw sensor telemetry, uncalibrated against a reference probe. CatBoost training-size analysis indicates that limited early data contributed to its poor performance.

\textbf{Future work.} Despite its small scale, this pilot demonstrates the potential of AI-based forecasting for soil moisture at multiple depths and local temperature. As data accumulate across new sites and seasons, future work will investigate when fine-tuning TimesFM provides meaningful gains over zero-shot forecasting, how much data site-specific models require, and whether forecasting knowledge can transfer to newly instrumented sites. We will also systematically evaluate the efficiency of different forecasting approaches, including inference time, memory footprint, and training cost, to understand the trade-offs and identify appropriate model choices for different deployment settings.

\begin{ack}
This work was supported by a seed funding grant from the Environmental Informatics Hub, Monash University, awarded to Mohammad Goudarzi and Vanessa Wong.
\end{ack}

\section*{References}
\small

[1] Hadi, H., Ray, D.K., Borrelli, P., Gerber, J.S., Shokri, N., Roman, S., \& Wuepper, D. (2026). Land degradation is associated with larger crop yield gaps across global croplands. \textit{Nature Food}, 7, 574--584.

[2] Wong, V.N.L., Greene, R.S.B., Dalal, R.C., \& Murphy, B.W. (2010). Soil carbon dynamics in saline and sodic soils: A review. \textit{Soil Use and Management}, 26(1), 2--11.

[3] Vereecken, H., Huisman, J.A., Pachepsky, Y., Montzka, C., van der Kruk, J., Bogena, H., et al. (2014). On the value of soil moisture measurements in vadose zone hydrology: A review. \textit{Water Resources Research}, 50(2), 787--834.

[4] Norman, L.M., Lal, R., Wohl, E., Fairfax, E., Gellis, A.C., \& Pollock, M.M. (2022). Natural infrastructure in dryland streams (NIDS) can establish regenerative wetland sinks that reverse desertification and strengthen climate resilience. \textit{Science of The Total Environment}, 849, 157779.

[5] Vereecken, H., Schnepf, A., Hopmans, J.W., Javaux, M., Or, D., Roose, T., Vanderborght, J., et al. (2016). Modeling soil processes: Review, key challenges, and new perspectives. \textit{Vadose Zone Journal}, 15(5), 1--20.

[6] Fang, K., Shen, C., Kifer, D., \& Yang, X. (2017). Prolongation of SMAP to spatiotemporally seamless coverage of continental U.S. using a deep learning neural network. \textit{Geophysical Research Letters}, 44(21), 11030--11039.

[7] Yang, G., Chang, J., Zhang, L., Kam, J., Zheng, Y., Wang, Y., Guo, A., Hussain, A., \& Liu, A. (2026). Harnessing deep learning for dual gains in S2S-scale soil moisture forecasting and flash drought mechanisms. \textit{Water Resources Research}, 62(3), e2025WR041466.

[8] Das, A., Kong, W., Sen, R., \& Zhou, Y. (2024). A decoder-only foundation model for time-series forecasting. \textit{ICML 2024}. arXiv:2403.14735.

[9] Garza, A., \& Mergenthaler-Canseco, M. (2023). TimeGPT-1. arXiv:2310.03589.

[10] Ansari, A.F., Stella, L., Turkmen, C., et al. (2024). Chronos: Learning the language of time series. arXiv:2403.07815.

[11] Shi, X., Wang, S., Nie, Y., Li, D., Ye, Z., Wen, Q., \& Jin, M. (2025). Time-MoE: Billion-scale time series foundation models with mixture of experts. \textit{ICLR 2025}. arXiv:2409.16040.

[12] Zhang, P., Qin, H., Zhang, H., Guo, Z., Wang, G., \& Bi, J. (2026). Zero-shot transfer capabilities of the Sundial foundation model for leaf area index forecasting. \textit{AAAI 2026 Workshop on AI in Agriculture (AgriAI)}. arXiv:2511.20004.

[13] Vaswani, A., Shazeer, N., Parmar, N., Uszkoreit, J., Jones, L., Gomez, A.N., Kaiser, {\L}., \& Polosukhin, I. (2017). Attention is all you need. \textit{NeurIPS 2017}.

[14] Nie, Y., Nguyen, N.H., Sinthong, P., \& Kalagnanam, J. (2023). A time series is worth 64 words: Long-term forecasting with transformers. \textit{ICLR 2023}.

[15] Prokhorenkova, L., Gusev, G., Vorobev, A., Dorogush, A.V., \& Gulin, A. (2018). CatBoost: unbiased boosting with categorical features. \textit{NeurIPS 2018}.

[16] Zeng, A., Chen, M., Zhang, L., \& Xu, Q. (2023). Are transformers effective for time series forecasting? \textit{AAAI 2023}.

[17] Zhang, Y., \& Gilpin, W. (2025). Zero-shot forecasting of chaotic systems. \textit{ICLR 2025}. arXiv:2408.06543.

\end{document}